\documentclass{article}
\usepackage{spconf,amsmath,amssymb,amsfonts,amsxtra,bm}
\usepackage{microtype}
\usepackage{mathtools}
\usepackage{booktabs}
\usepackage[utf8]{inputenc}
\usepackage[T1]{fontenc}
\usepackage{amsthm}
\usepackage[numbers]{natbib}
\usepackage{graphicx}
\usepackage{subcaption}
\usepackage{algorithm2e}
\usepackage{natbib}
\usepackage{xspace}
\usepackage{paralist}
\usepackage{enumerate}
\usepackage{enumitem}
\usepackage{hyperref}
\usepackage{url}
\usepackage{colortbl}
\usepackage{makecell,multirow}       %
\usepackage{nicefrac}       %
\usepackage{thmtools}  
\usepackage{xspace}
\usepackage{footnote, tablefootnote}
\usepackage{float}
\floatstyle{plaintop}
\restylefloat{table}
\usepackage{epstopdf}
\usepackage{latexsym}
\usepackage{bbm}
\usepackage{cleveref}
\usepackage{algorithmic}
\usepackage{hyperref}
\newcommand{\leqnomode}{\tagsleft@true}
\newcommand{\reqnomode}{\tagsleft@false}

\newcommand{\Cc}{\mathcal{C}}

\newcommand{\Sc}{\mathcal{S}}

\newcommand{\E}{\mathbb{E}}

\newcommand{\reals}{\mathbb{R}}

\makeatother

\makeatletter
\def\thanks#1{\protected@xdef\@thanks{\@thanks
        \protect\footnotetext{#1}}}
\makeatother

\usepackage[
  color=blue!20!white,
  textsize=tiny,
  colorinlistoftodos,
  prependcaption,
]{todonotes}

\newtheorem{theorem}{Theorem}

\newtheorem{Assumption}{Assumption}

\newtheorem{lemma}{Lemma}

\theoremstyle{definition}

\newtheorem{remark}{Remark}

\clearpage{}%

\title{Joint Unsupervised and Supervised Training for\\ Automatic Speech Recognition via Bilevel Optimization}%

\name{
A F M Saif $^\dag$, Xiaodong Cui$^{\star}$, Han Shen$^\dag$, Songtao Lu$^{\star}$, Brian Kingsbury$^{\star}$, Tianyi Chen$^\dag$ \thanks{This work was  supported by the Rensselaer-IBM AI Research Collaboration (\url{http://airc.rpi.edu}), part of the IBM AI Horizons Network (\url{http://ibm.biz/AIHorizons}.}}
\address{$^\dag$Rensselaer Polytechnic Institute, 
Troy, NY, USA\\
 $^{\star}$IBM Research AI, T. J. Watson Research Center, 
	Yorktown Heights, NY, USA}
\begin{document}
\ninept
\maketitle
\begin{abstract}
In this paper, we present a novel bilevel optimization-based training approach to training acoustic models for automatic speech recognition (ASR) tasks that we term {bi-level joint unsupervised and supervised training (BL-JUST)}.  {BL-JUST employs a lower and upper level optimization with an unsupervised loss and a supervised loss respectively, leveraging recent advances in penalty-based bilevel optimization to solve this challenging ASR problem with affordable complexity and rigorous convergence guarantees.} To evaluate BL-JUST, extensive experiments on the LibriSpeech and TED-LIUM v2 datasets have been conducted. BL-JUST achieves superior performance over the commonly used pre-training followed by fine-tuning strategy.  

\end{abstract}
\section{Introduction}
\label{sec:intro}

Automatic Speech Recognition (ASR) is a popular research area that plays a vital role in improving both human–human and human–machine communication~\cite{hinton2012deep}. It enables the smooth conversion of speech signals into written text. Deep neural networks (DNNs) have been used to improve ASR performance~\cite{hinton2012deep}. However, their performance usually relies on a large amount of labeled data that is expensive to obtain \cite{deng2013new}. To overcome this issue, the two-stage approach of pre-training followed by fine-tuning that we term PT+FT thereafter has been actively studied and yielded good performance \cite{baevski2019vq,baevski2020wav2vec,hsu2021hubert,chiu2022self}. In this strategy, a DNN model is first trained in an unsupervised fashion on a large amount of unlabeled data. It is then fine-tuned in a supervised training fashion on a small amount of labeled data in downstream applications. 


In the PT+FT approach, the ASR model is pre-trained independently without considering any feedback from downstream fine-tuning tasks~\cite{rosenstein2005transfer}.  Consequently, the fine-tuning step has limited control over the upstream pre-training. Hence, when the pre-trained and fine-tuned domains are not closely related, there is a mismatch in transferring knowledge. In some cases, it may even adversely impact the model's performance, a phenomenon referred to as \emph{negative transfer}~\cite{pan2009survey,wang2019characterizing}. Furthermore, the PT+FT approach necessitates two separate training loops, where the first loop is for pre-training and the second loop is for fine-tuning. This disconnected training increases the complexity and time of training. To overcome the known limitation of the two-stage PT+FT approach, this paper proposes a new joint unsupervised and supervised training approach for ASR that we term Bi-Level Joint Unsupervised and Supervised Training (BL-JUST).  Different from a related work \cite{bai2022joint} on joint unsupervised and supervised training, BL-JUST allows the use of distinct datasets for PT and FT stages and proposes a recursive training method based on the bilevel optimization technique which has seen increasing success in a broad variety of applications~\cite{liu2021investigating,crockett2022bilevel,chen2023learning,franceschi2018bilevel,finn2017model}.



In general, bilevel optimization problems are optimization problems where the feasible set is determined (in part) using the solution set of a second optimization problem~\cite{dempe2020bilevel}. Determining the feasible set is generally called the lower-level problem and the second parametric optimization problem is called the upper-level problem~\cite{vicente1994bilevel,sabach2017first}. 

In the context of ASR, we regard the unsupervised training stage, which has the goal of learning generic representations of speech signals that can be fine-tuned for a particular task, as the lower-level problem. Ideally, the result of this lower-level problem is a set of initial model parameters or weights of backbone layers that promote successful and efficient learning in the upper-level supervised training, which minimizes a task-specific loss given the lower-level parameters.

Our contributions in this paper can be summarized as follows:
\begin{itemize}
\itemsep -0.05em 
    \item[\bf C1)] To overcome the data scarcity and negative transfer, we formulate the joint unsupervised and supervised training of acoustic models in ASR tasks as a bilevel optimization problem. 
    \item[\bf C2)] Leveraging recent advances in penalty-based bilevel optimization, we propose BL-JUST, a novel joint unsupervised and supervised training approach to solving the resultant bilevel problem with a rigorous convergence guarantee.
    \item[\bf C3)] We carry out extensive experiments on the Librispeech and TED-LIUM v2 datasets, showning that BL-JUST has superior performance over the PT+FT approach in terms of both accuracy and runtime.  
\end{itemize}

\section{Problem Formulation}
\label{sec:method}
We provide preliminaries on bilevel optimization, and formulate ASR acoustic model training as a bilevel optimization problem.
 
\subsection{Bilevel optimization preliminaries}
 

Bilevel optimization is a two-level optimization problem. The upper-level problem attempts to optimize an objective function while being constrained by factors influenced by the solutions of the lower-level problem.  
 If the upper-level objective is defined as $f:\mathbb{R}^{d_\phi} \times \reals^{d_\theta}\mapsto \reals$ and the lower-level objective is defined as $g:\mathbb{R}^{d_\phi} \times \reals^{d_\theta}\mapsto \reals$, then the bilevel optimization problem can be written as
\begin{equation}
    \min_{\phi\in\mathbb{R}^{d_\phi},\theta\in\reals^{d_\theta}} f(\phi,\theta)\,\,\,\,\,\,\,  \text{s.t.}\,\,\,\, \theta \in \mathcal{S}(\phi) \coloneqq \arg\min_{\theta\in\reals^{d_\theta}} g(\phi,\theta)
\end{equation}
where $\Sc(\phi)$  are non-empty and closed sets given any $\phi\in \mathbb{R}^{d_\phi}$. 
Though bilevel optimization has a wide range of applications, it is difficult to solve due to its non-convex and non-differentiable nature~\cite{brotcorne2020special}. 
Recently, some implicit gradient-based and unrolled differentiation-based methods have been developed to solve bilevel optimization problems; see e.g.,~\cite{pedregosa2016hyperparameter,chen2022single,coupled2023bilevel}. However, those methods are costly and thus are not scalable to large models used in ASR.

    \vspace{-0.2cm}
\subsection{Bilevel optimization for acoustic model training}

To reformulate the acoustic model training  as a bilevel optimization problem, we first introduce the unsupervised and supervised objective functions we will use in this work.

For unsupervised training, we use the InfoNCE loss~\cite{oord2018representation} to learn a good representation of the input speech from unlabeled data. Given a set of $N$ samples $X\coloneqq\{x_1, x_2, ..., x_N\}$ and a similarity metric $f(\cdot, \cdot)$, the InfoNCE loss function, defined as
\begin{equation}
\label{eq:nce}
L_{\rm NCE}(\theta)=-\E\left[ \log \dfrac{f\left( x_{t+p},C_{t}(\theta)\right) }{\sum _{x'\in X'}f\left( x',C_{t}(\theta)\right) }\right]
\end{equation}
aims to maximize the probability of predicting the future sample $x_{t+p}$ given a contextual representation $C_t(\theta)$ generated from the speech sequence $\{x_1, x_2, ..., x_t\}$ up to time $t$ using a neural network parameterized by $\theta$. 
In the InfoNCE loss \eqref{eq:nce}, $x_{t+p}$ and $C_t(\theta)$ form a positive sample pair, and samples from the speech sequence at other time steps, denoted as $x' \in X'$, together with $C_t(\theta)$ form negative pairs.

\begin{figure}[tbp]
    \centering
    \includegraphics[width=0.985\linewidth]{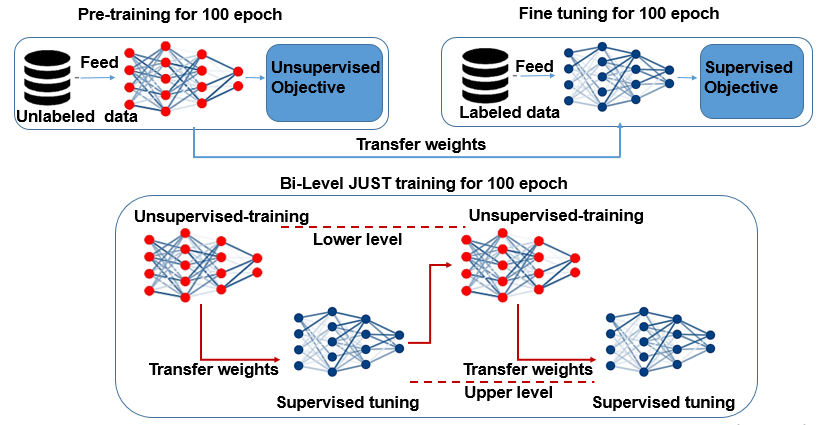}
    \caption{Comparison between the proposed BL-JUST training method (bottom) with the PT+FT method (upper).}
    \label{fig:BL-JUST}
    \vspace{-0.2cm}
\end{figure}

For supervised training, the Connectionist Temporal Classification (CTC) loss~\cite{graves2006connectionist} is used. 
When the input sequence is $x_n$ and the label sequence is $y_n$, the CTC objective minimizes the negative log-likelihood of the label sequence $y_n$, given by  
\begin{equation}
\label{ctc}
    L_{\rm CTC}(\phi,\theta) = \frac{1}{N}\sum_{n=1}^{N}-\log P(y_{n}|z(x_{n};\phi, \theta))
\end{equation}
where $z(x_{n};\phi, \theta)$ is the output of the model, $\phi$ is the parameters of the model's classification layer, and $\theta$, which we call the ``backbone,'' includes all the parameters except those from the classification layer. 

In BL-JUST, we combine the two objective functions into a bilevel optimization problem, where the upper-level objective is the CTC loss and the lower-level objective is the InfoNCE loss:
\begin{equation}\label{eq:original prob}
\begin{aligned}
& \underset{\phi,\theta}{\min}
&& L_{\rm CTC}(\phi,\theta) \\
& \text{s.t.}
&& \theta \in\Sc\coloneqq \arg\min_{\theta} L_{\rm NCE}(\theta).
\end{aligned}
\end{equation}
In \eqref{eq:original prob}, the lower-level unsupervised training problem serves as the constraint for the backbone model parameters $\theta$ in the optimization of the upper-level supervised objective.

The rationale for using the above bilevel optimization formulation \eqref{eq:original prob} is that due to the overparameterization of ASR models, while there might be multiple values of  $\theta$ that minimize the unsupervised InfoNCE loss, we want to find the one that also optimizes the supervised CTC loss.  
By solving the above bilevel optimization problem \eqref{eq:original prob}, we can use the supervised objective to guide the unsupervised training, and find a better feature representation with improved ASR performance.

\section{Joint Unsupervised and Supervised Training}

 This section presents the algorithm and analyzes its convergence. 
  \vspace{-0.2cm}
\subsection{Training}

\SetNlSty{textbf}{(}{)} 
\begin{algorithm}[t]
\caption{The BL-JUST Training Algorithm}
\label{algo:BL-JUST}
\SetAlgoLined

\SetKwInOut{Input}{Input}
\SetKwInOut{Output}{Output}

\Input{Labeled data $(x,y)$, unlabeled data $(x_{t+p},x')$, learning rates $\alpha$ and $\beta$, and penalty constant $\gamma$}

\SetKw{KwTo}{to}
\SetKw{KwDownTo}{downto}
\SetKw{KwAnd}{and}
\SetKw{KwOr}{or}
\SetKw{KwTrue}{true}
\SetKw{KwFalse}{false}

Randomly initialize $\theta_1, \phi_1$ \\
\For{$k=1, 2, 3, \ldots, K$}{
    Update: $\theta_{k+1}$ via \eqref{update_theta}\\
    Update $\phi_{k+1}$ via \eqref{update_phi}
}
\Output{$\theta_{K}$, $\phi_{K}$} \medskip
\end{algorithm}
 
To solve the joint training problem \eqref{eq:original prob} efficiently, we employ the penalty-based reformulation of the bilevel problem in  \eqref{eq:original prob}; that is
\begin{equation}\label{eq:penalized prob}
\min_{\phi,\theta} ~F_\gamma(\phi,\theta) \coloneqq \, L_{\rm CTC}(\phi,\theta) + \gamma L_{\rm NCE}(\theta)
\end{equation}
where $\gamma>0$ is a penalty constant that will be specified later. We will rigorously establish the equivalence of the penalized reformulation \eqref{eq:penalized prob} and the original bilevel problem \eqref{eq:original prob} in the next subsection. 

With the penalized reformulation \eqref{eq:penalized prob}, BL-JUST jointly optimizes the unsupervised and supervised training in a single loop.  At first, the backbone parameters $\theta$ are randomly initialized with $\theta_1$ and the classification head parameters $\phi$ are initialized with $\phi_1$. Then applying the gradient descent to the penalized formulation \eqref{eq:penalized prob} with respect to $\theta$, we update the backbone model parameters $\theta$; that is, the $k$th iteration follows
\begin{equation}
\label{update_theta}
    \theta_{k+1} = \theta_k - \alpha \nabla_\theta L_{\rm CTC}(\phi_k,\theta_k) - \alpha\gamma \nabla_\theta L_{\rm NCE}(\theta_k)
\end{equation}
where $\alpha\!>\!0$ is the learning rate. Please note that, different from the conventional multi-objective training by linearly combining two objective functions, the two gradient terms in \eqref{update_theta} are evaluated using labeled and unlabeled data, respectively. 

 Similarly, we can update the classification head parameters $\phi$ as 
\begin{equation}
\label{update_phi}
    \phi_{k+1} = \phi_k - \beta \nabla_{\phi} L_{\rm CTC}(\phi_k,\theta_k)
\end{equation}
where $\beta\!>\!0$ is a pre-defined learning rate. Hence, the unsupervised training is now coupled with the supervised training step. The BL-JUST algorithm is summarized in Algorithm \ref{algo:BL-JUST}.  

\begin{remark}[Two-stage training versus BL-JUST training]
   The process of PT+FT in a two-stage training model can be seen in Figure \ref{fig:BL-JUST}. During pre-training, the model is trained for a number of iterations, and then fine-tuned in a separate loop. However, the fine-tuning has no impact on the pre-training. In contrast, the BL-JUST method alternates between unsupervised and supervised training, allowing the unsupervised training to receive feedback from the supervised training. This leads to a more cohesive and effective training process.
\end{remark}

\subsection{Convergence}



Before performing the convergence analysis, we first make the following assumptions.
\begin{Assumption}\label{assumption: QGEB}
    Consider the following assumptions:
    \begin{enumerate}[label=(\alph*)]
    \itemsep -0.1em 
        \item The loss $L_{\rm CTC}(\cdot,\theta)$ is $L$-Lipschitz continuous in $\phi$ given  $\theta$.
        \item Define $L_{\rm NCE}^*=\min_\theta L_{\rm NCE}(\theta)$. There exists $\mu>0$ such that 
    $L_{\rm NCE}(\theta)$ satisfies the PL inequality
    $\|\nabla L_{\rm NCE}(\theta)\|^2 \geq \frac{1}{\mu}\big(L_{\rm NCE}(\theta)-L_{\rm NCE}^*\big)$.
    \item The gradient $\nabla F_\gamma(\phi,\theta)$ is $L_\gamma$-Lipschitz continuous. 
    \end{enumerate}
\end{Assumption}
The Lipschitz gradient assumptions are standard in finite-time convergence analysis of gradient based methods, see e.g.,~\cite{bottou2018optimization}.
Recent studies found that over-parameterized neural networks can lead to losses that satisfy the PL inequality~\citep{liu2022loss}.

Following~\cite[Proposition 2]{shen2023penalty}, we obtain the next lemma.
\begin{lemma}[Equivalence of the penalized formulation]\label{lemma:solution} 
Under Assumption \ref{assumption: QGEB}, with a prescribed accuracy $\delta>0$, set $\gamma \geq L\sqrt{3\mu\delta^{-1}}$.
If $(x_\gamma,y_\gamma)$ is a local/global solution of \eqref{eq:penalized prob}, it is also a local/global solution of the following approximate problem of \eqref{eq:original prob} with some $\epsilon_\gamma \leq \delta$:
\begin{equation}\label{eq:approx original prob}
  \underset{\phi\in\Cc,\theta\in\Theta}{\min}
  L_{\rm CTC}(\phi,\theta)\,\,\,\, \,\,\text{s.t.}\,\,\,\,
  L_{\rm NCE}(\theta)-\min_{\theta} L_{\rm NCE}(\theta) \leq \epsilon_\gamma.
\end{equation}
\end{lemma}
This lemma suggests one can solve the penalized problem in \eqref{eq:penalized prob} locally/globally to solve the original problem in \eqref{eq:original prob}. To solve \eqref{eq:penalized prob}, we can view the BL-JUST algorithm in Algorithm \ref{algo:BL-JUST} as performing the projected gradient descent method.
Next, we present the theorem on the convergence of Algorithm \ref{algo:BL-JUST} based on~\cite[Theorem 3]{shen2023penalty}.
\begin{theorem}[Convergence rate of BL-JUST]\label{the:nonconvex xy convergence}
Consider Algorithm \ref{algo:BL-JUST}. 
Suppose Assumptions \ref{assumption: QGEB} holds.
Select an accuracy $\delta$ and
$\beta\in (0, L_\gamma^{-1}],~\gamma \text{ chosen by Lemma \ref{lemma:solution}}$. Then it holds that

i) With $C\!=\!\inf_{(\phi,\theta)} L_{\rm CTC}(\phi,\theta)$, it holds that
\begin{equation}
    \frac{1}{K}\sum_{k=1}^K \|\nabla F_\gamma(\phi_k,\theta_k)\|^2 \leq \frac{18\big(F_\gamma(\phi_1,\theta_1)-C\big)}{\beta K}+\frac{10 L^2 L_\gamma^2}{K}.\nonumber
\end{equation}
ii) Suppose $\lim_{k\rightarrow\infty}(\phi_k,\theta_k)=(\phi^*,\theta^*)$, then $(\phi^*,\theta^*)$ is a stationary point of \eqref{eq:penalized prob}. If $(\phi^*,\theta^*)$ is a local/global solution of the penalized problem \eqref{eq:penalized prob}, then it is a local/global solution of \eqref{eq:approx original prob} with some $\epsilon_\gamma\leq\delta$.
\end{theorem}
Theorem \ref{the:nonconvex xy convergence} suggests the iteration complexity to achieve $\epsilon$-stationary point 
 is $\mathcal{O}(L_\gamma\epsilon^{-1})$, which matches the complexity of the gradient descent-based supervised training method~\cite{bottou2018optimization}.

\section{Experiments}

In this section, we carry out experiments on ASR tasks to show the effectiveness of BL-JUST and compare it with supervised baselines and the commonly used PT+FT strategy. Code is available at \href{https://github.com/afmsaif/Joint-unsupervised-and-supervised-training-for-automatic-speech-recognition-via-bilevel-optimization}{GitHub}.


 \begin{table}[t]
\centering

\caption{WERs under various hyperparameter settings of BL-JUST using conformer and CNN-LSTM acoustic models on Librispeech. There are 860 hours of data in the lower-level unsupervised training and 100 hours of data in the upper-level supervised training.}

\label{tab:grid_search_results}
\begin{tabular}{@{}cccrr@{}} \toprule
& $\alpha$ & $\beta$ & {\bf Rate of} $\gamma$ & {\bf WER} \\ \midrule
\multirow{4}{*}{Conformer}& $5\!\times\!10^{-2}$ & $5\!\times\!10^{-2}$ & 0.01 & 11.2\% \\
& $5\!\times\!10^{-3}$ & $5\!\times\!10^{-3}$  & 0.004 & 10.0\% \\
& $5\!\times\!10^{-4}$ & $5\!\times\!10^{-3}$  & 0.002 & 9.2\% \\
& \textbf{$5\!\times\!10^{-3}$ } & \textbf{$5\!\times\!10^{-4}$} & \textbf{0.002} & \textbf{8.3\%} \\ \midrule
\multirow{4}{*}{CNN-LSTM}&$5\!\times\!10^{-2}$ & $5\!\times\!10^{-2}$ & 0.01 & 14.0\% \\
&$5\!\times\!10^{-3}$ & $5\!\times\!10^{-3}$  & 0.004 & 11.5\%\\
&$5\!\times\!10^{-4}$ & $5\!\times\!10^{-3}$  & 0.002 & 10.6\%\\
&\textbf{$5\!\times\!10^{-3}$} & \textbf{$5\!\times\!10^{-4}$} & \textbf{0.002} & \textbf{9.4\%}\\ \bottomrule
\end{tabular}
\end{table}
\begin{table}[tb]
    \centering
    \begin{tabular}{@{}llrr@{}} \toprule
    {\bf Model} & {\bf Method} & {\bf Test-clean} & {\bf Test-other} \\ \midrule
    Conformer & Supervised baseline & 11.0\% & 23.5\% \\
              & PT+FT               & 10.3\% & 18.6\% \\
              & BL-JUST                &  9.2\% & 16.4\% \\ \midrule
    CNN-LSTM  & Supervised baseline & 16.9\% & 25.4\% \\
              & PT+FT   & 14.0\% & 19.6\% \\
              & BL-JUST                & 10.6\% & 18.7\% \\ \bottomrule
    \end{tabular}

    \caption{WERs of supervised training baseline, PT+FT and BL-JUST using conformer and CNN-LSTM acoustic models, respectively, on Librispeech. There are 100 hours of speech in training.}
     \label{tab:tab1}
\end{table}
 
\subsection{Experimental Setting}

\noindent\textbf{Dataset} \ \  We evaluate BL-JUST on LibriSpeech~\cite{panayotov2015librispeech} and TED-LIUM  v2~\cite{rousseau2014enhancing}. The Librispeech dataset has 960 hours of speech. The TED-LIUM v2 dataset has 207 hours of speech for training and about 4 hours for validation. The sampling rate of both datasets is 16KHz. 

 \vspace{0.1cm}
\noindent\textbf{Model}  \ \  We use two model architectures for acoustic models.  One is conformer~\cite{gulati2020conformer}, and the other is a bi-directional LSTM model on top of residual convolutional layers, which is referred to as CNN-LSTM. There are 7 conformer blocks in the conformer model with 512 hidden units and 8 64-dimensional attention heads in each conformer block. The convolution kernel size is 31. The conformer model has about 52M parameters. The CNN-LSTM model has 3 convolutional layers on top of which are 5 LSTM layers. There are 32 feature maps in each convolutional layer with a kernel size 3x3 and stride 1. There are 256x2 hidden units in each LSTM layer. The LSTM acoustic model has about 27M parameters. For both acoustic models, the input is 80-dimensional logmel features and the output is 1,000 BPE units. {We used beamsearch decoder and WordPiece for tokenization.}   

 \vspace{0.1cm}
\noindent\textbf{Training strategy} \ \  All models are trained using the AdamW optimizer \cite{loshchilov2017decoupled}. For supervised models, the learning rate starts with $5\!\times\!10^{-4}$. In the PT+FT and BL-JUST, learning rates start from $5\!\times\!10^{-3}$ for unsupervised training and $5\!\times\!10^{-4}$ for supervised training. A learning rate scheduler was employed to reduce the learning rate when the validation loss plateaued. Specifically, we set the patience value to 20 and the reduction factor to 0.1. Consequently, our training process monitored the validation loss for 20 epochs, and if no improvement was observed, it reduced the learning rate by a factor of 0.1. We trained  BL-JUST and the baseline methods for 100 epochs. In the case of the PT+FT method, we first pre-trained the model for 100 epochs and subsequently fine-tuned it for an additional 100 epochs. In BL-JUST, $\gamma$ is monotonically increasing from 0 with a linear rate of 0.002 over epochs. These hyper-parameters are optimized based on the ASR performance in Table \ref{tab:grid_search_results}. SpecAug \cite{park2019specaugment} is used for data augmentation in the training.



\begin{table}[tb]
    \centering
    \begin{tabular}{@{}llrr@{}} \toprule
    {\bf Model} & {\bf Method} & {\bf Test-clean} & {\bf Test-other} \\ \midrule
    Conformer & Supervised baseline & 8.6\% & 11.3\% \\
              & PT+FT   & 6.5\% & 9.2\% \\
              & BL-JUST                & 5.2\% & 7.8\% \\ \midrule
    CNN-LSTM  & Supervised baseline & 9.9\% & 16.5\% \\
              & PT+FT               & 8.5\% & 14.0\% \\
              & BL-JUST                & 6.2\% & 13.2\% \\ \bottomrule
    \end{tabular}

    \caption{WERs of supervised training baseline, PT+FT and BL-JUST using conformer and CNN-LSTM acoustic models, respectively, on Librispeech. There are 960 hours of speech in training.}
    
     \label{tab:tab2}
\end{table}
\subsection{ASR Performance}

We compared BL-JUST with two training approaches. We first compare BL-JUST with supervised training where the acoustic models are trained with labeled data in the conventional manner, which is referred to as the supervised baseline.  We also compare BL-JUST with the PT+FT strategy where the acoustic model is first trained with unlabelled data in an unsupervised way and then it is fine-tuned with labeled data under supervised training. 

\begin{table}[t]
    \centering
    \begin{tabular}{@{}llrr@{}} \toprule
    {\bf Model} & {\bf Method} & {\bf Test-clean} & {\bf Test-other} \\ \midrule
    Conformer & Supervised baseline & 11.0\% & 23.5\% \\
              & PT+FT   &  9.8\% & 14.5\% \\
              & BL-JUST    &  8.3\% & 13.2\% \\ \midrule
    CNN-LSTM      & Supervised baseline & 16.9\% & 25.4\% \\
              & PT+FT   & 13.2\% & 15.7\% \\
              & BL-JUST    &  9.4\% & 14.5\% \\ \bottomrule
    \end{tabular}

    \caption{WERs of supervised training, PT+FT and BL-JUST using conformer and CNN-LSTM acoustic models, respectively, on Librispeech. There are 860 hours of speech in lower-level unsupervised training and 100 hours of speech in upper-level supervised training.}
     \label{tab:tab3}
\end{table}

We consider three setups on Librispeech with a 100-hour/860-hour split.  In Setup 1, we used only 100 hours of speech to train supervised baseline, PT+FT and BL-JUST. For PT+FT and BL-JUST, we used the same 100 hours of data for both unsupervised and supervised training; see the results in Table \ref{tab:tab1}. In Setup 2, we used all 960 hours of speech to re-run the unsupervised and supervised training like Setup 1; see the results in Table \ref{tab:tab2}. In Setup 3, we use 100 hours of speech data for a supervised baseline. In PT+FT and BL-JUST, we use 860 hours of speech for unsupervised training and 100 hours of speech for supervised training; see the results in Table \ref{tab:tab3}.  Analogously, we conduct experiments on TED-LIUM v2 using the conformer acoustic model. We use $70\%$ of its data for unsupervised training and $30\%$ for supervised training; see the results in Table \ref{tab:ted}. 

From the tables, BL-JUST outperforms the conventional PT+FT and the supervised baseline that only relies on available labeled data. Specifically, in Table \ref{tab:tab1} BL-JUST outperforms PT+FT by almost $1\%$ using conformer and $3.5\%$ using CNN-LSTM. with more data for unsupervised and supervised training this margin increases. In Tables \ref{tab:tab2} and \ref{tab:tab3}, BL-JUST outperforms PT+FT by almost $1.3\%$ and $1.5\%$ using conformer and $2.3\%$ and $3.8\%$ using CNN-LSTM, respectively. Similar observations can also be made in Table \ref{tab:ted} on TED-LIUM v2.

\begin{figure}[hb]
    \centering
    \includegraphics[width=0.75\linewidth]{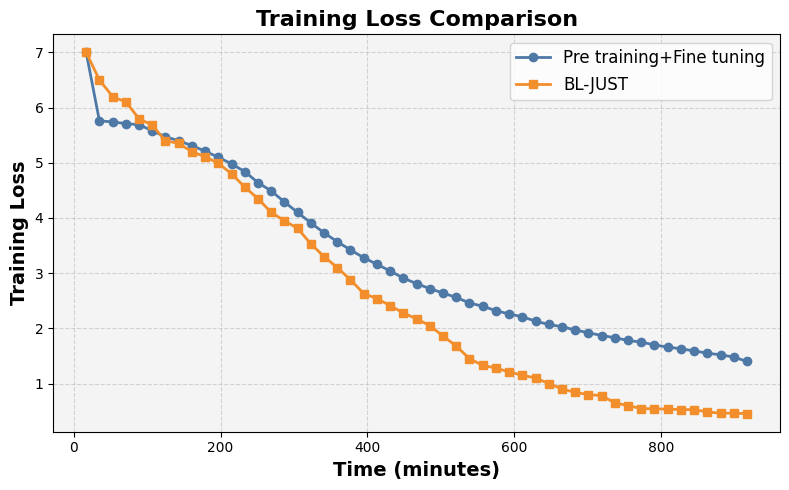}
    \caption{Training losses of BL-JUST vs. PT+FT on 100 hours of speech in Librispeech. The acoustic model is conformer.}
    \label{fig:losstime}
\end{figure}

Note that  BL-JUST combines unsupervised and supervised training in a single training loop which effectively reduces the number of epochs needed. Figure \ref{fig:losstime} shows a comparison between the computational cost of PT+FT and BL-JUST. To generate Figure \ref{fig:losstime}, we first recorded the time taken by the PT+FT method for pre-training and separately for fine-tuning. We then combined these times to calculate the total time. For the BL-JUST method, we only needed to measure a single training time since BL-JUST combines both supervised and unsupervised training within a single loop. The loss value here present is the supervised CTC loss. It shows that, BL-JUST takes much less time to achieve a certain loss value compared to PT+FT.

\begin{table}[tb]
    \centering
\begin{tabular}{@{}lrrr@{}} \toprule
    {\bf Method} & {\bf Valid. set} & {\bf Test set} & {\bf \% of Labeled data} \\ \midrule
    Supervised baseline   &  9.6\% &  7.0\% & $100\%$ \\
    Supervised baseline  & 25.4\% & 21.3\% &  $30\%$ \\
    PT+FT        & 15.2\% & 13.4\% & $30\%$ \\
    BL-JUST         & 14.4\% & 12.2\% &  $30\%$ \\ \bottomrule
\end{tabular}
\caption{WERs on TED-LIUM v2 using a conformer acoustic model.}
\label{tab:ted}
\end{table}
\begin{table}[t]
  \centering
  \caption{Effect of rate of change of penalty constant in bilevel optimization under various amounts of supervised and unsupervised data.}
  \label{tab:penalty-constant}
  \begin{tabular}{cccc} \toprule
   \multirow{2}{*}{\textbf{$\gamma$ change rate}} & \multicolumn{3}{c}{\textbf{supervised/unsupervised}}  \\  \cline{2-4}
                                    &    100h/100h    &    100h/860h    &    960h/960h     \\ \midrule
    0.000 &  13.6\% &  13.2\% &  12.1\% \\                  
    0.010 &  12.4\% &  11.8\% &  10.9\% \\                  
    0.008 &  11.8\% &  11.3\% &  10.2\% \\                  
    0.006 &  11.0\% &  10.4\% &  \hphantom{0}9.0\% \\                  
    0.004 &  10.8\% &  \hphantom{0}9.6\% &  \hphantom{0}8.5\% \\                  
    0.002 &  \hphantom{0}9.2\% &  \hphantom{0}8.3\% &  \hphantom{0}7.8\% \\  \bottomrule
  \end{tabular}
\end{table}
 

\vspace{-0.2cm}
\subsection{Effect of penalty constant}
We investigate the impact of the penalty constant, $\gamma$, of bilevel optimization, to BL-JUST.  Table \ref{tab:penalty-constant} shows WERs under different growth rates of $\gamma$. We gradually increase the growth rate starting from $0$. A slowly increasing $\gamma$ gives better performance over the constant $\gamma$. The best rate is $0.002$ which is consistent across all  setups on Librispeech.


 \section{Conclusions}
In this work, we proposed BL-JUST, an effective training strategy that performs unsupervised and supervised training jointly based on a principled bilevel optimization framework. BL-JUST introduces a feedback loop between unsupervised and supervised training compared to the conventional two-stage PT+FT training strategy. Extensive experiments on Librispeech and TED-LIUM v2 datasets show that BL-JUST can give superior performance in reducing WERs and improving training efficiency.


{\fontsize{8.1}{9.95}\selectfont
\bibliographystyle{plainnat}
\bibliography{refs}
}

\end{document}